\documentclass[manuscript,screen]{acmart}
\AtBeginDocument{%
  }

\setcopyright{acmlicensed}
\copyrightyear{2018}
\acmYear{2018}
\acmDOI{XXXXXXX.XXXXXXX}
\acmConference[Conference acronym 'XX]{Make sure to enter the correct
  conference title from your rights confirmation email}{June 03--05,
  2018}{Woodstock, NY}
\acmISBN{978-1-4503-XXXX-X/2018/06}

\usepackage{subcaption}
\usepackage{algorithm}
\usepackage{algorithmic}
\usepackage{multirow}
\usepackage{longtable}
\begin{document}

\title{Mitigating Individual Skin Tone Bias in Skin Lesion Classification through Distribution-Aware Reweighting}


\author{Kuniko Paxton}
\email{k.azuma-2021@hull.ac.uk}
\orcid{0009-0000-8897-9775}
\authornotemark[1]
\affiliation{%
  \institution{University of Hull}
  \city{Hull}
  \state{East Riding of Yorkshire}
  \country{United Kingdom}
}

\author{Zeinab Dehghani}
\email{Z.Dehghani@hull.ac.uk}
\orcid{0009-0007-4617-9455}
\authornotemark[1]
\affiliation{%
  \institution{University of Hull}
  \city{Hull}
  \state{East Riding of Yorkshire}
  \country{United Kingdom}
}

\author{Koorosh Aslansefat}
\email{K.Aslansefat@hull.ac.uk}
\orcid{0000-0001-9318-8177}
\affiliation{%
  \institution{University of Hull}
  \city{Hull}
  \state{East Riding of Yorkshire}
  \country{United Kingdom}
}

\author{Dhavalkumar Thakker}
\email{D.Thakker@hull.ac.uk}
\orcid{0000-0003-4479-3500}
\affiliation{%
  \institution{University of Hull}
  \city{Hull}
  \state{East Riding of Yorkshire}
  \country{United Kingdom}
}

\author{Yiannis Papadopoulos}
\email{Y.I.Papadopoulos@hull.ac.uk}
\orcid{0000-0001-7007-5153}
\affiliation{%
  \institution{University of Hull}
  \city{Hull}
  \state{East Riding of Yorkshire}
  \country{United Kingdom}
}


\begin{abstract}
Skin color has historically been a focal point of discrimination, yet fairness research in machine learning for medical imaging often relies on coarse subgroup categories, overlooking individual-level variations. Such group-based approaches risk obscuring biases faced by outliers within subgroups.  This study introduces a distribution-based framework for evaluating and mitigating individual fairness in skin lesion classification.  We treat skin tone as a continuous attribute rather than a categorical label, and employ kernel density estimation (KDE) to model its distribution. We further compare twelve statistical distance metrics to quantify disparities between skin tone distributions and propose a distance-based reweighting (DRW) loss function to correct underrepresentation in minority tones. Experiments across CNN and Transformer models demonstrate: (i) the limitations of categorical reweighting in capturing individual-level disparities, and (ii) the superior performance of distribution-based reweighting, particularly with Fidelity Similarity (FS), Wasserstein Distance (WD), Hellinger Metric (HM), and Harmonic Mean Similarity (HS).  These findings establish a robust methodology for advancing fairness at individual level in dermatological AI systems, and highlight broader implications for sensitive continuous attributes in medical image analysis.
\end{abstract}



\received{31 July 2025}

\maketitle

\section{Introduction}
The deployment of advanced machine learning (ML) systems in healthcare has amplified the urgency of ensuring that automated predictions are not only accurate but also equitable. Among the pillars of responsible AI, safety, robustness, transparency, and accountability, fairness occupies a critical position, especially as regulatory frameworks such as the EU AI Act and the U.S. AI Bill of Rights increasingly demand guarantees against discriminatory outcomes \cite{Hofl2025,eu2024,wh2022}. In dermatology, where automated diagnostic systems are rapidly entering clinical workflows, fairness considerations are particularly salient because skin color has historically been a focal point of social discrimination and remains a legally protected attribute \cite{jablonski2021skin}. Unlike other sensitive attributes, skin color presents unique methodological challenges for fairness research. It is multidimensional, continuous, and often subjectively perceived, making categorical assignments, such as ethnicity labels or Fitzpatrick skin types, both inconsistent and reductive\cite{groh2022towards, barocas2023fairness, krishnapriya2021analysis, heldreth2024skin, barrett2023skin}.  Moreover, skin tone distributions in dermatology datasets are highly imbalanced: lighter tones are typically overrepresented, while darker tones are scarce, reflecting both epidemiological trends and historical biases in data collection   \cite{paxton2025evaluating, rezk2022improving, wen2022characteristics, li2021estimating, ferrara2023fairness, guo2022bias, daneshjou2022disparities, seth2024does}. As a result, performance disparities have repeatedly been reported, with classifiers achieving higher accuracy for lighter skin tones and reduced reliability for darker tones  \cite{lin2024preserving, muthukumar2019color, buolamwini2018gender, bevan2022detecting, pakzad2022circle, sarridis2023towards, kinyanjui2019estimating}. 

Existing mitigation strategies, such as data augmentation through generative models \cite{shaham2019singan,kalb2023towards,pakzad2022circle,pakzad2024addressing,choi2018stargan,rezk2022leveraging,tanantong2024improving,alaluf2022third,munia2025dermdiff}, adversarial learning \cite{li2021estimating,hwang2020exploiting,kshettry2024enhancing,correa2023adversarial,du2022fairdisco}, or counterfactual and pruning approaches \cite{wu2022fairprune,chiu2023toward,kong2024achieving,meyer2022fair,bayasi2025biaspruner,ghadiri2024xtranprune,xue2024bmft} have improved fairness at the group level but remain tied to categorical assumptions. These methods often overlook subtle intra-group nuances, require extensive computational resources, and may not guarantee that generated or reweighted samples reflect clinically valid tonal variations. Recent efforts \cite{paxton2025evaluating,pundhir2024biasing} have highlighted the potential of treating skin tone as a continuous attribute. These group-based representations risk obscuring within-group disparities and leave individuals whose tones diverge from group norms disproportionately disadvantaged. Yet, evaluations have largely been limited to single statistical measures and have not comprehensively addressed mitigation.

This study advances the field by introducing a distribution-aware framework for both measuring and mitigating individual-level fairness in skin lesion classification. Specifically, we model skin tone as a continuous distribution of Individual Typology Angle (ITA) values and employ kernel density estimation (KDE) to capture tonal nuances across datasets. We systematically evaluate twelve statistical distance metrics to quantify disparities between distributions and propose a Distance-based Reweighting (DRW) method that assigns loss weights inversely proportional to skin tone density. By doing so, our approach addresses imbalances across the entire tonal spectrum without relying on arbitrary categorical divisions.

Through extensive experiments on CNN- and Transformer-based architectures, we investigate four central research questions: (i) whether fairness analyses based on categorical types can consistently detect unfairness across individuals; (ii) whether correlations exist between skin tone representation and classification accuracy; (iii) whether reweighting strategies grounded in continuous distributions can mitigate underrepresentation effects; and (iv) which statistical metrics are most effective for capturing skin tone nuances. By answering these questions, we provide both a methodological contribution to fairness research in continuous attributes and practical insights for developing equitable dermatological AI systems.

Our contributions are fourfold:
\begin{enumerate}
    \item \textbf{Continuous skin tone modelling}: We treat skin tone as a distribution of Individual Typology Angle (ITA) values rather than a categorical attribute, capturing its full variability using kernel density estimation (KDE).
    \item \textbf{Systematic evaluation of statistical metrics}: We conduct a comprehensive comparison of eleven distance and similarity functions—including Fidelity Similarity, Wasserstein Distance, Hellinger Metric, and Harmonic Mean Similarity—to determine their suitability for fairness assessment in continuous attributes.
    \item \textbf{Bias mitigation through Distance-based Reweighting (DRW)}: We propose a novel reweighting method that adjusts the loss function based on skin tone density, thereby correcting underrepresentation across the tonal spectrum without synthetic augmentation.
    \item \textbf{Extensive empirical validation}: We evaluate our approach on three widely used architectures (ResNet50, MobileNetV2, and Vision Transformer) under 72 experimental conditions, comparing baselines, categorical reweighting, DRW, and hybrid methods to rigorously assess performance and fairness outcomes.
    
\end{enumerate}
The remainder of this paper is structured as follows. Section 2 reviews prior work on fairness in skin lesion classification. Section 3 presents our methodology for continuous skin tone distribution modelling and bias mitigation. Section 4 details the experimental setup, while Section 5 reports results across categorical and individual fairness analyses. Section 6 discusses the findings in relation to our research questions, architectural sensitivity, limitations, and future work. Finally, Section 7 concludes by highlighting the broader implications of distribution-aware fairness for dermatological AI and medical imaging more generally.

\section{Related Work}
In previous studies, a significant theme in efforts to mitigate bias against skin tone in skin lesion classification has been how to prevent performance degradation for darker skin tones, which are in the minority, and overcome the instability associated with this. In this section, we focus on methods to handle minority skin color data and provide an overview of the diversity measures that have been proposed.

First, the most commonly proposed methods in recent years were data augmentation and generation techniques. These have become feasible due to improvements in image generation techniques. For example, \cite{ali2024web} converted images to the HSV colour space and generated 180 color variations from a single image, enabling learning across diverse skin tones for the same lesion structure. Although 180 color patterns are significant, the justification for the selected colors was not argued. In addition, this method faced practical challenges of increased loading and computational costs. \cite{yuan2022edgemixup} proposed a mix-up technique that linearly combined edge images and original images to emphasize lesions while reducing the influence of skin color. Nevertheless, this method did not insist on removing the skin tone bias as it prioritized lesion emphasis over fairness. \cite{corbin2023assessing} proposed a practical color conversion algorithm that adjusts the L and b components of the LAB colour space based on the ITA value, replacing high-cost methods such as StarGAN. This manipulated pixel values according to the ITA difference between Fitzpatrick skin types, enabling realistic skin color conversion. CIRCLe \cite{pakzad2022circle} used StarGAN \cite{choi2018stargan} to generate images of different skin types and minimized the difference between the feature vectors of the generated images and the original images by regularizing the loss, thereby achieving robust feature learning for skin color changes. \cite{rezk2022improving,rezk2022leveraging} synthesized images containing lesions and dark-skinned style images, and generated new dark-skinned lesion images by minimizing style loss and content loss based on features extracted by the Visual Geometry Group network \cite{simonyan2014very}. \cite{abhari2023mitigating} used CycleGAN \cite{zhu2017unpaired}, an image generation model that converts one domain into another to expand the minority skin color dataset. They applied it to a skin cancer classification model. Using StyleGAN3, \cite{tanantong2024improving} artificially generated images with dark skin tones and contributed to a skin lesion classification model specifically for Thailand. These methods were effective in supplementing minority skin color samples. However, neither method focused on subtle differences, such as nuances within skin color groups, and there is no guarantee that the generated colors are appropriate as necessary skin colors. Skin tone diversity is not limited to differences in lightness or between groups, but also includes nuances in the generated colors themselves, which tend to be neglected. These methods require large computational resources, too.

The second type of method is the Unlearning and Adversarial Learning Method. \cite{bevan2022detecting} applied Turning a Blind Eye (TABE) \cite{alvi2018turning} and Learning Not to Learn (LNTL) \cite{kim2019learning}, to mitigate biases caused by skin color. TABE stood out in that disparities related to skin tone were reduced. TABE consisted of classification loss for the main task, bias classification loss for sensitive attributes, and confusion loss to suppress bias classification performance. In the case of \cite{li2021estimating}, they introduced the Feature Generator, Bias Discriminator, and Fairness Critic, along with the primary task classifier. \cite{hwang2020exploiting} also added a sensitive attribute classifier encoder and formed a feature structure based on the sensitive attributes of the original image as an anchor using Feature Independency Triplet Loss. FairDisCo \cite{du2022fairdisco} is a complex method consisting of four components: target branch, sensitive attribute branch, contrast branch, and feature extractor. Methods based on unlearning or adversarial learning adjust the loss function so that 'the classification of the main task is maintained while features related to skin color are not learned.’ This requires the assumption that ‘skin color attributes can be classified,’ in other words, label information is mandatory. Therefore, these methods are dependent on the framework of group fairness and cannot guarantee continuous nuances in skin tone and fairness at the individual level.

In addition, methods such as counterfactual method \cite{dash2022evaluating}, prune learning \cite{wu2022fairprune,chiu2023toward,kong2024achieving}, federated learning \cite{xu2022achieving,xing2025achieving}, and distillation learning \cite{pundhir2024biasing} have been proposed to reduce skin dependency. These methods also assumed that skin color was categorized, and did not sufficiently consider individual skin color nuances. To the best of our knowledge, the example of quantitative measurement of skin nuances in previous studies is that of \cite{paxton2024measuring, paxton2025evaluating}. However, their verification was limited to measurements using the Wasserstein Distance, which was based solely on skin color differences between arbitrary samples, and no evaluation based on the representative distribution of the entire training data was conducted. Therefore, in this study, we measured skin tone nuances using 11 distribution distance functions and applied them to three types of models (2 CNN-based models and 1 Transformer-based model) to comprehensively verify the most appropriate measurement method. Additionally, we mitigated bias through a non-complex approach of extending categorical reweighting to continuous values and validated the results from both group and individual fairness perspectives. The findings of this study provide new directions for ensuring fairness in skin color, a unique and sensitive continuous attribute.

\subsection{Research Questions}
\begin{enumerate}
    \item Can the fairness analysis based on skin types consistently detect unfairness across diverse individual skin colors?
    \item Are there any correlations between the number of data instances in diverse skin colors and accuracy? 
    \item If there are correlations between them, can reweighting using these relationships address the bias caused by the minority of skin tone nuances? How effective is it?
    \item Which method is the most suitable to measure skin tone to identify and mitigate the skin color bias?
\end{enumerate}

\subsection{Main Contributions}
\begin{enumerate}
    \item Addressing \textbf{RQ1–RQ3}, we introduced a novel technique to evaluate the accuracy performance of ultimate individual fairness by utilizing kernel density estimation and mitigating it using a re-weighting loss function.
    \item For \textbf{RQ1–RQ3}, we conducted a comprehensive comparison of twelve statistical distance functions to quantify disparities between continuous distributions of skin tone.
    \item In response to \textbf{RQ4}, we determined the most suitable distance metric for measuring skin color based on the comprehensive comparison above.
\end{enumerate}
We contribute to improving fairness for individuals within a group in terms of skin color in the skin lesion classification task, as evidenced by the outcomes above.

\section{Methodology}
To address fairness in skin lesion classification, we operationalize skin tone as a \textbf{continuous distribution} rather than a categorical attribute. Our methodology has two main stages: (i) \textbf{data analysis}, where we extract and model skin tone distributions, and (ii) \textbf{bias mitigation}, where we introduce a distribution-aware reweighting strategy.

\subsection{Data Analysis}
\label{lab:data_analysis}

\subsubsection{Skin Pixels Extraction}
Dermoscopic images often include artefacts such as hair that obscure skin regions. Following common preprocessing practice\cite{Kasmi2023SharpRazor,suiccmez2023detection,abbas2011hair,khan2021classification,kaur2022hairlines,talavera2020hair,bardou2022hair,jaworek2013hair,ramella2021hair}, we applied a pipeline of morphological transformations—including opening, Black Hat filtering\cite{soille1999morphological}, and Contrast-Limited Adaptive Histogram Equalisation \cite{zuiderveld1994contrast}, to detect and mask hair. Lesion regions were further excluded using publicly available segmentation masks, ensuring that only pure skin regions were analysed.

\subsubsection{Skin Tone Representation with ITA}
\label{sec:distance}
We adopt the \textbf{Individual Typology Angle (ITA)} as a well-established quantitative representation of skin tone. After converting images into CIELab color space using \textit{scikit-image} \cite{skimage}, ITA values were computed per pixel according to Equation~\ref{eq:ita}, where \(L^*\) denotes lightness and \(b^*\) the blue–yellow axis.  
\begin{equation}
\label{eq:ita}
    \mathrm{ITA}(L^*, b^*) = \frac{180}{\pi}\arctan\!\Big(\frac{L^* - 50}{b^*}\Big)\in[\theta_{\min},\theta_{\max}].
\end{equation}
The resulting ITA values constitute a distribution for each image, spanning the minimum $\theta_{\min}$ to maximum $\theta_{\max}$ angles observed in the training set (with bin width of \(1^\circ\)). This approach mirrors prior uses of ITA in dermatological imaging to reflect continuous skin color variation \cite{kinyanjui2019estimating,osto2022individual}. 

For reference, we calculated the aggregated median skin tone distribution using the equation below. Given a set of the ITA distributions $A =\left\{ a_{i,1},..., a_{i,N} \right\}$ and $\text{bin}$ starts with $\theta_{\min}$ to $\theta_{\max}$, we converted $A$ to histograms with \ref{eq:hist}.
\begin{equation}
\label{eq:hist}
    q_{i,j} = \frac{1}{n_{\Delta}}\sum_{k=1}^{n_{i}}1\left\{ a_{i, k}\in [\text{bin}_{j-1}, \text{bin}) \right\}, j=1,...,K
\end{equation}
Lastly, we aggregated all training data histograms into a single histogram.
\begin{equation}
\label{eq:aggregation}
Q^{\circ}=\text{Median}_{i=1,...,N}q_{i,j}
\end{equation}

\subsubsection{Measuring Skin Color Nuance Differences}
\label{se:distance}
To capture differences between each image distribution and the aggregated reference , we evaluated twelve statistical distance and similarity measures (Table \ref{tab:metrics}), spanning information-theoretic, geometric, and similarity-based formulations. For each image, the distance was computed as:
\begin{equation}
\label{eq:skin-similarity}
D(x)=\mathrm{Distance}(Q_x,Q^{\circ})\in[0,1].
\end{equation}

Lower $D(x)$ indicate greater similarity to the reference distribution, while higher values correspond to rarer tones. This approach avoids the limitations of discrete skin type categories, enabling fine-grained detection of underrepresented tones.

\begin{table}[ht]
\centering
\caption{Distances applied in the experiments $\mathsf{D}(P,Q)$ used for comparing skin tone distributions. 
Each can be substituted into Equation \ref{eq:skin-similarity} without altering the rest of the framework.}
\label{tab:metrics}
\renewcommand{\arraystretch}{2} 
\setlength{\tabcolsep}{8pt} 
\begin{tabular}{lll}
\toprule
\textbf{Symbol} & \textbf{Name} & \textbf{Definition / Formula} \\ 
\midrule
$\mathrm{AD}$ & \textbf{Anderson--Darling Distance} & 
$\displaystyle \mathrm{AD}(P,Q) = \int \frac{(F_P(t) - F_Q(t))^2}{F_Q(t)(1-F_Q(t))}\,dt$ \\[6pt]

$\mathrm{CVM}$ & \textbf{Cramér--von Mises Distance} & 
$\displaystyle \mathrm{CVM}(P,Q) = \int (F_P(t) - F_Q(t))^2 dt$ \\[6pt]

$\mathrm{FS}$ & \textbf{Fidelity Similarity} (Bhattacharyya) & 
$\displaystyle \mathrm{FS}(P,Q) = \sum_i \sqrt{p_i q_i}$ \\[4pt]

$\mathrm{HS}$ & \textbf{Harmonic Mean Similarity} & 
$\displaystyle \mathrm{HS}(P,Q) = \sum_i \frac{2p_i q_i}{p_i + q_i}$ \\[6pt]

$\mathrm{HM}$ & \textbf{Hellinger Metric} & 
$\displaystyle \mathrm{HM}(P,Q) = \sqrt{1 - \sum_i \sqrt{p_i q_i}}$ \\[6pt]

$\mathrm{KL}$ & \textbf{Kullback--Leibler Divergence} & 
$\displaystyle \mathrm{KL}(P\|Q) = \sum_i p_i \log\frac{p_i}{q_i}$ \\[6pt]

$\mathrm{KS}$ & \textbf{Kolmogorov--Smirnov Distance} & 
$\displaystyle \mathrm{KS}(P,Q) = \sup_t |F_P(t) - F_Q(t)|$ \\[6pt]

$\mathrm{KP}$ & \textbf{Kuiper Distance} & 
$\displaystyle \mathrm{KP}(P,Q) = \sup_t [F_P(t) - F_Q(t)] + \sup_t [F_Q(t) - F_P(t)]$ \\[6pt]

$\mathrm{KD}$ & \textbf{Kruglov Distance} & 
$\displaystyle \mathrm{KD}(P,Q) = \sum_i \frac{|p_i - q_i|}{1 + |p_i - q_i|}$ \\[6pt]

$\mathrm{PF}$ & \textbf{Patrick–Fisher Distance} &
$\displaystyle \mathrm{PF}(P,Q) = \sqrt{\sum_i (p_i - q_i)^2}$ \\

$\mathrm{WD}$ & \textbf{Wasserstein Distance (1-D)} & 
$\displaystyle \mathrm{WD}(P,Q) = \int |F_P(t) - F_Q(t)|\,dt$ \\[6pt]

\bottomrule
\end{tabular}
\end{table}

The comparison of those statistical distance functions aims to select optimal metrics for measuring skin color differences and mitigating bias caused by data density. As explained above, each distance function enables the capture of differing characteristics in relation to skin tone nuance. Therefore, it is possible to clarify which measurement is most suitable for optimizing the density distribution of data instances based on skin tone by comparing these methods. This means that it is necessary to select a distance function that can more accurately capture the nuances of skin color.

In this step, we demonstrate disparities between skin color types, which are commonly used in traditional studies on unfairness mitigation, and our approach, which treats skin tone in a continuous numerical manner. In other words, we reveal latent skin tone bias in the categorical skin types. Although there are several skin color scales, we selected the Fitzpatrick Skin Color Scale \cite{fitzpatrick1988validity} because it is the most frequently employed and provides a clear threshold score.

\subsubsection{Performance Evaluation}
Fairness metrics are often used in the AI fairness studies, such as Equal Opportunity \cite{hardt2016equality} and Equalized Odds \cite{hardt2016equality}, which are required for their respective groups. However, our approach aims to detect individual unfairness. Thus, we focus on accuracy, that is, whether the prediction is correct or not, for our analytics prospects. Then, we investigated the correlation between the data sample size and accuracy in the measured skin color diversity space. Whether dark or light, extreme skin tones become outliers in their number and follow a normal distribution, so the correlation is assumed not to be linear. Therefore, Spearman's correlation coefficient is selected. Specifically, each distance was normalized to the range of 0 to 1, and then divided into equal bins based on the distribution of the training data. The test data were separated using the exact bin boundaries. We then calculated the Spearman correlation coefficient between the average classification accuracy of the test data in each bin and the number of samples in the corresponding bin in the training data. This evaluation method is also used for the validity evaluation of our fairness reweighting loss function.

\subsection{Bias Mitigation Algorithm}
\label{lab:bias_mitigation}

\begin{figure}
    \centering
    \includegraphics[width=1\linewidth]{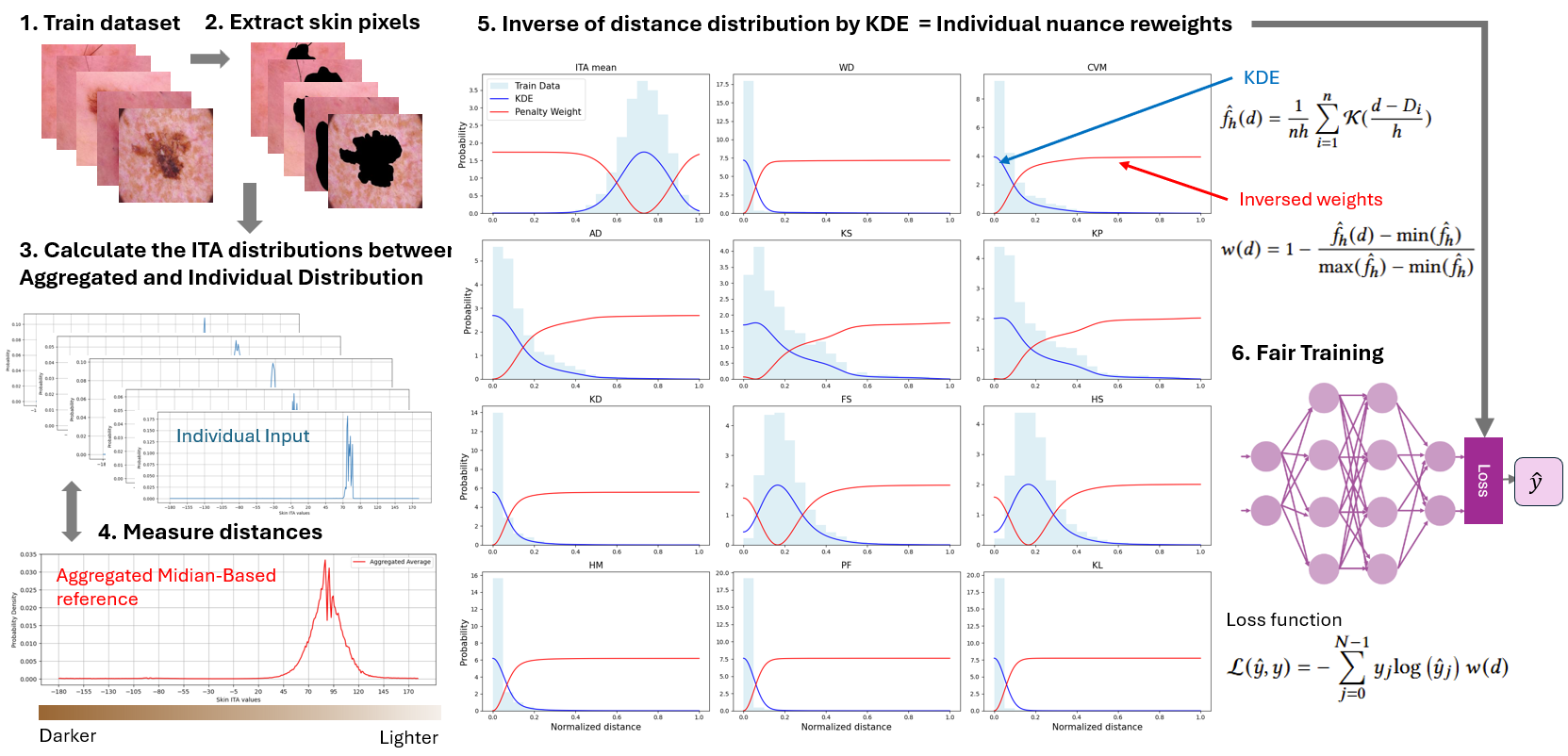}
    \caption{Fair Training Framework Procedure with Distance Metrics for Experiments: First, only skin regions are extracted from the training data, and skin color is quantified using ITA to obtain its distribution. Subsequently, the distance to the reference skin tone distribution, which is the aggregated median, is calculated, and based on the distance, inverse weighting is applied to the loss function while keeping the values continuous. This is a very simple method that extends the distribution-based reweighting technique traditionally used for categorical data to continuous data.}
    \label{fig:fair_training}
\end{figure}

Fig. \ref{fig:fair_training} overviews its bias mitigation framework. The details are explained in each subsection. The performance bias, hypothetically, is caused by the lack of data on minority skin tones in the training process. To correct this imbalance of skin tone, we estimate the probability density over the skin tone diversity space using kernel density estimation (KDE). The distances in the dataset, observed using one of the metrics in the subsection, are denoted by $D$, and the number of data points is $n$. Let $\mathcal{K}$ be the kernel function, which we selected to follow a Gaussian distribution, and $h$ is the bandwidth parameter. Using Eq.~\ref{eq:kde}, which refers to \cite{silverman2018density}, we estimate the kernel density at a given distance $d$ from the input $x$ to predict. 

\begin{equation}
\label{eq:kde}
\hat{f}_{h}(d) = \frac{1}{nh}\sum_{i=1}^{n}\mathcal{K}(\frac{d-D_{i}}{h})
\end{equation}

For inverse density weighting, we simply subtract the calculated density from 1 to assign higher weights to inputs associated with lower density regions. To ensure that the weights remain within an appropriate range, we normalized the KDE as follows Eq.\ref{eq:weight}

\begin{equation}
\label{eq:weight}
w(d) = 1 - \frac{\hat{f}_{h}(d)-\text{min}(\hat{f}_{h})}{\text{max}(\hat{f}_{h})-\text{min}(\hat{f}_{h})}
\end{equation}

Since we handle multi-class classification in this project, we apply cross-entropy loss in the same way as the general cases using Eq.~\ref{eq:cross_entropy}, where the equation refers to \cite{chaudhury2024math}. $N$ denotes the number of classes and $y$ and $\hat{y}$ are respectively ground truth and predictions. Then, the weights calculated by Eq.~\ref{eq:weight} are applied. The detailed procedure is described in Algorithm~\ref{alg:alg_lossfunction}.

\begin{equation}
\label{eq:cross_entropy}
    \mathcal{L}(\hat{y}, y) = -\sum_{j=0}^{N-1}y_{j}\text{log}\left( \hat{y}_{j} \right) w(d)
\end{equation}

\begin{algorithm}
\caption{Fair Cross Entropy Loss Function}
\label{alg:alg_lossfunction}
\begin{algorithmic} 
\STATE \textbf{Input}: Prediction $\hat{y}$, Ground truth $y$, Distance $d$, Class Weights $c$ \\
\STATE \textbf{Output}: Loss $l$
\STATE Initialize list of loss $l$
\FOR {$n$ in ${N}$}
\STATE Calculate cross-entropy loss and append to $l_{n}$
\IF {$c_{n}$ is set}
\STATE Apply class weights $c_{n}$ to $l_{n}$
\ENDIF
\STATE Normalize $d$ following the range of the distance of the training dataset
\STATE Compute KDE $\hat{f}_{h}(d_{n})$ of $d_{n}$ by Eq.~\ref{eq:kde}
\STATE Compute weight $w(d_{n})$ from $\hat{f}_{h}(d_{n})$ by~\ref{eq:weight}
\STATE Apply distance weight $w(d_{n})$ by Eq.\ref{eq:cross_entropy} to $l_{n}$
\STATE Append $l_n$ to $l$
\ENDFOR
\STATE Compute final loss with $\text{Weighted Mean}(l)$
\end{algorithmic}
\end{algorithm}

To validate our approach, we implemented a categorical reweighting algorithm as proposed by \cite{kamiran2012data}. This algorithm is commonly used, and this method was expanded to a study conducted by \cite{du2022fairdisco} against skin color bias and class imbalance. FairDisco \cite{du2022fairdisco} shared their implementation through the website; hence, we followed them (Eq. \ref{eq:reweighing}). 

\begin{equation}
\label{eq:reweighing}
    w_{r}(x) = \frac{P_{\text{exp}}(x(s)=s_{i},x(y)=y_{j})}{P_{\text{obs}}(x(s)=s_{i},x(y)=y_{j})}
\end{equation}, where $S$, $P_{\text{exp}}$, and $P_{\text{obs}}$ are denoted as skin color types, the joint probability distribution of skin color types and class labels, and the expected joint probability. The equation refers to \cite{du2022fairdisco}.

Then, we analyzed whether their categorical reweighting algorithm covers individual fairness for skin tone. Moreover, we also implemented combined methods of categorical and our approach. We calculated categorical and class weights using Eq.~\ref{eq:reweighing}, and then applied Eq.~\ref{eq:re_cross_entropy} with each of our distance methods in Table \ref{tab:metrics}.

\begin{equation}
\label{eq:re_cross_entropy}
\mathcal{L}(\hat{y}, y) = -\sum_{j=0}^{N-1}y_{j}\text{log}\left( \hat{y}_{j} \right) w_{r}(d)w(d)
\end{equation}

\section{Experimental Setup}
\textbf{Dataset:} The Human Against Machine 10000 training images dataset (HAM) \cite{tschandl2018ham10000, tschandl2020human} was selected as its skin lesion segmentation data is publicly available. The HAM has seven categories and is frequently used in studies on skin color bias mitigation.
\textbf{Models: }In this study, we used 50-layer Residual Networks(ResNet) \cite{he2016deep}, MobileNetV2 (MobileNet) \cite{sandler2018mobilenetv2}, and Vision Transformer-B16 (ViT) \cite{dosovitskiy2020image} with backbones that were pre-trained using ImageNet \cite{deng2009imagenet}. They were fine-tuned with the following settings, and the general performance is in Table 1. The highest F1-score models were selected as the vanilla models. These hyper-parameters were applied to ensure consistency across fairness model trainings, including our individual and conventional categorical reweighting experiments.

\section{Result}
\subsection{Performance in Skin Color Types}
\label{sec:categorical}
Table \ref{tab:performance_skin_type} shows the accuracy and F1 score for each skin type in the baseline model (BL) and the conventional categorical reweighting method (CARW). In the BL, the accuracy for skin types 2 to 4 consistently exceeds 90\%, while type 6 is significantly lower, and type 5 has high accuracy and F1 scores. Type 1 is around the overall average. These trends are consistent with the biased distribution of data sizes, where types 2–4 have relatively more data and stable prediction, while types 5 and 6 have less data and greater variability. After CARW, types 2 and 3 maintained high performance, and optimization to reduce inter-category gaps improved types 5 and 6 compared to the BL, but gaps remained, and overall behavior was similar to the baseline.

\begin{table}[]
\centering
\caption{Performance of each Skin Color Type of Baseline and Categorical Reweighting Method}
\label{tab:performance_skin_type}
\resizebox{\textwidth}{!}{
\begin{tabular}{cr|rrrrrr|rrrrrr}
\hline
\multirow{3}{*}{Tone} & \multicolumn{1}{c|}{\multirow{3}{*}{Data}} & \multicolumn{6}{c|}{BL}                                                                                                                                                                  & \multicolumn{6}{c}{Reweighting}                                                                                                                                                         \\
                             & \multicolumn{1}{c|}{}                      & \multicolumn{3}{c}{ACC}                                                                    & \multicolumn{3}{c|}{F1-score}                                                               & \multicolumn{3}{c}{ACC}                                                                    & \multicolumn{3}{c}{F1-score}                                                               \\
                             & \multicolumn{1}{c|}{}                      & \multicolumn{1}{c}{ResNet} & \multicolumn{1}{c}{MobileNet} & \multicolumn{1}{c}{ViT} & \multicolumn{1}{c}{ResNet} & \multicolumn{1}{c}{Mobile} & \multicolumn{1}{c|}{ViT} & \multicolumn{1}{c}{ResNet} & \multicolumn{1}{c}{Mobile} & \multicolumn{1}{c}{ViT} & \multicolumn{1}{c}{ResNet} & \multicolumn{1}{c}{Mobile} & \multicolumn{1}{c}{ViT} \\ \hline
1 & 1585  & 0.837  & 0.840 & 0.846  & 0.765 & 0.767  & 0.749  & 0.842 & 0.840  & 0.836  & 0.738 & 0.738 & 0.718  \\
2 & 223  & 0.928  & 0.919 & 0.883  & 0.743 & 0.718  & 0.608  & 0.928 & 0.910  & 0.901  & 0.721 & 0.680 & 0.649  \\
3 & 100  & 0.900  & 0.930 & 0.950  & 0.454 & 0.593  & 0.676   & 0.940 & 0.910  & 0.930  & 0.823 & 0.566  & 0.735 \\
4 & 56   & 0.911   & 0.929  & 0.946 & 0.878 & 0.850 & 0.927  & 0.875 & 0.911 & 0.946 & 0.773 & 0.711  & 0.927  \\
5 & 12  & 0.917 & 1.000 & 1.000  & 0.667 & 1.000 & 1.000 & 0.917 & 1.000 & 1.000 & 0.933 & 1.000 & 1.000  \\
6  & 4 & 0.250 & 0.500 & 0.250  & 0.133 & 0.389 & 0.200  & 0.750 & 0.250 & 0.500  & 0.556 & 0.167 & 0.333  \\ \hline
Mean  & 330 & 0.790 & 0.853 & 0.813 & 0.606 & 0.719 & 0.693 & 0.875 & 0.804  & 0.852 & 0.757 & 0.644 & 0.727 \\ \hline \hline
\end{tabular}
}
\end{table}

\subsection{Performance Within Major Skin Color Type}
\label{sec:tone1}
\begin{figure}
    \centering
    \includegraphics[width=1\linewidth]{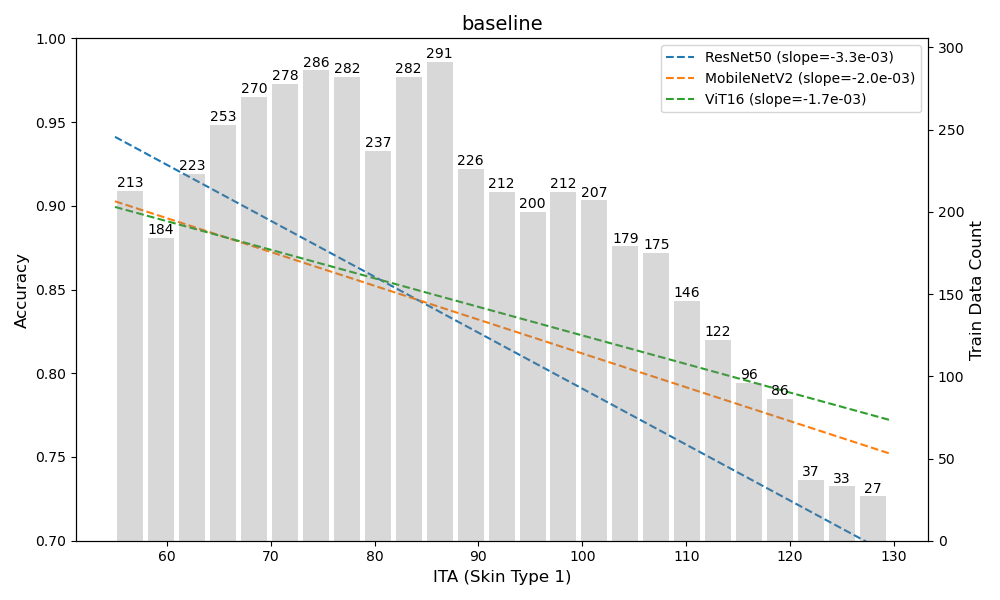}
    \Description{Description of the image for accessibility.}
    \caption{Performance and the Number of Data Differences in Skin Color Types 1 (Baseline)}
    \label{fig:tone1_bl}
\end{figure}

\begin{figure}
    \centering
    \includegraphics[width=1\linewidth]{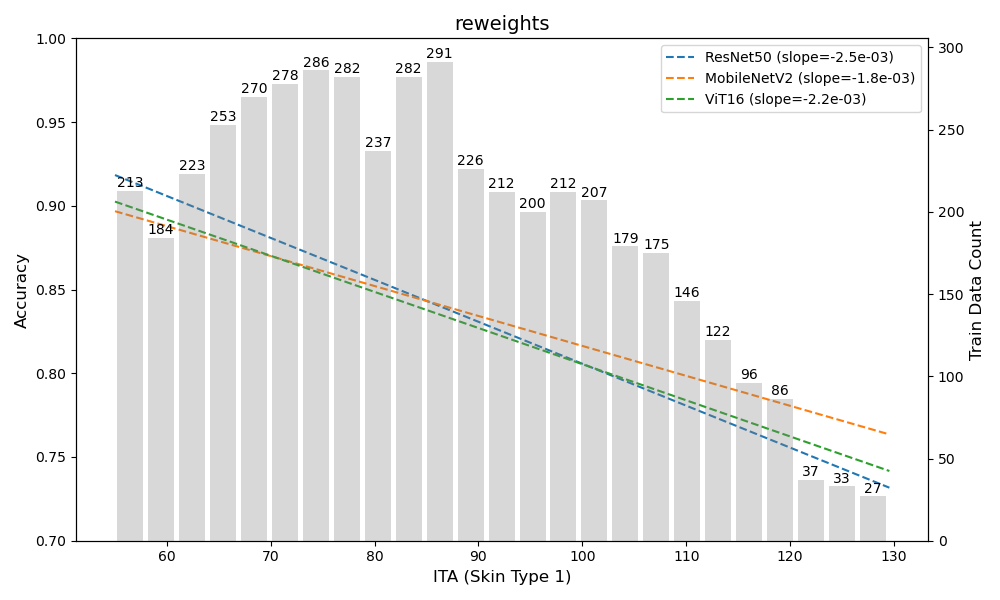}
    \Description{Description of the image for accessibility.}
    \caption{Performance and the Number of Data Differences in Skin Color Types 1 (Categorical Reweighting Method)}
    \label{fig:tone1_reweights}
\end{figure}

Figures \ref{fig:tone1_bl} and \ref{fig:tone1_reweights} show the relationship between accuracy and training data size for skin type 1 ($\text{ITA} \ge  55$), which has average accuracy and relatively high F1 scores, in the BL and CARW. Within type 1, there were also majority and minority, and the amount of training data gradually decreases as ITA exceeds 90°. The green, blue, and orange lines describe the correlations between accuracy and ITA scores using logistic regression. The slopes of the regression suggest decreases in accuracy as the number of data decreases. In other words, performance decreases due to insufficient samples in the minority range covered in the majority range. This implies that methods that detect skin color bias only at the category level may be insufficient. CARW showed a slight improvement over BL in ResNet and MobileNet, but the overall behavior was similar to BL.

\subsection{Correlation with the Number of Training Data}
\label{sec:correlation}

\begin{figure}
    \centering
    \includegraphics[width=1\linewidth]{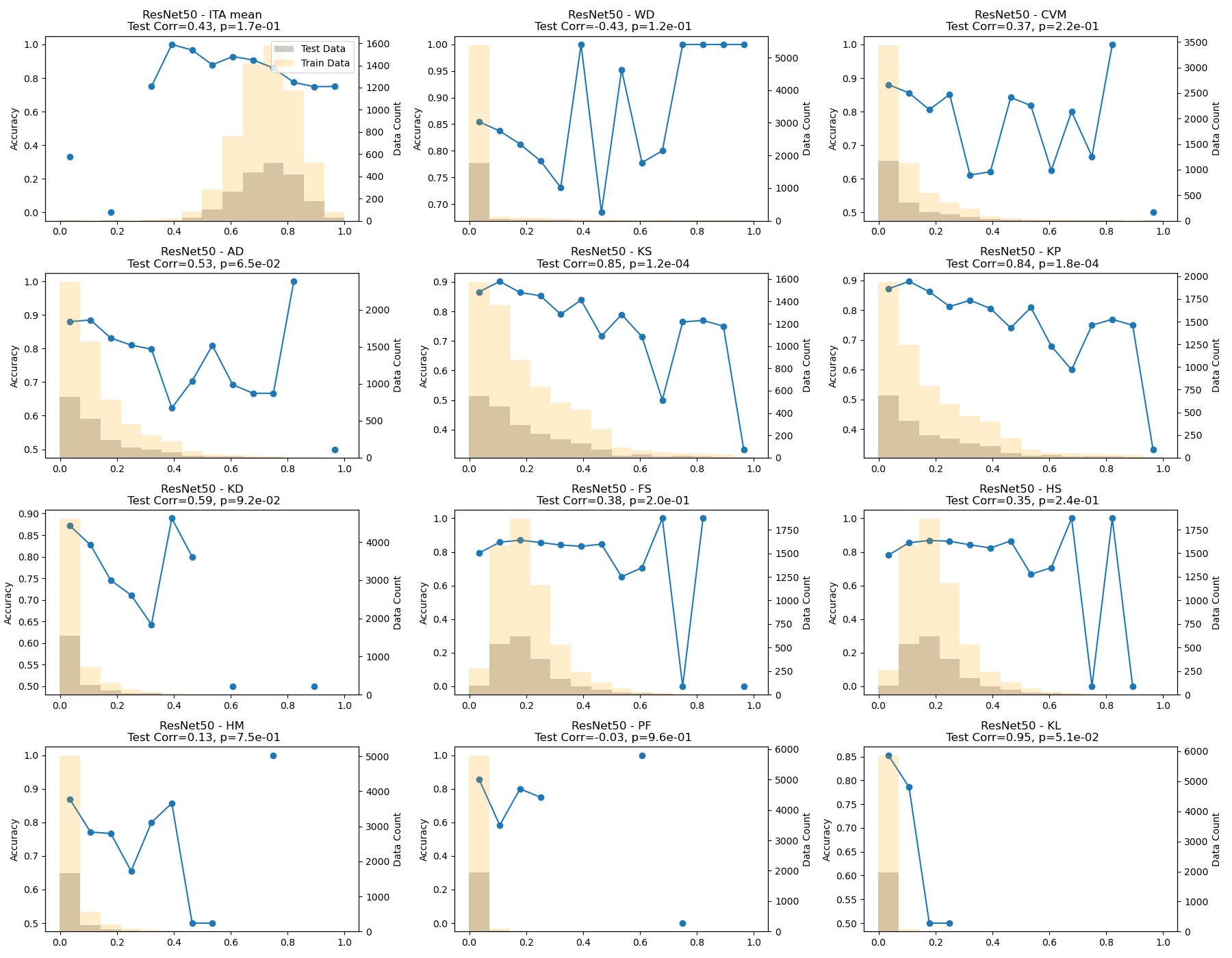}
    \Description{Description of the image for accessibility.}
    \caption{Correlation between Performance and the Number of Training Data (ResNet)}
    \label{fig:resnet_individual_correlation}
\end{figure}

Figure \ref{fig:resnet_individual_correlation} visualizes the correlation between skin tone nuances (11 distance metrics or ITA median) and accuracy in the ResNet baseline, together with the amount of training data (results for MobileNet and ViT-16 are shown respectively in the appendices \ref{fig:mobilenet_individual_correlation} and 2 \ref{fig:vit_individual_correlation}). The number of bins was set based on Sturges' rule for the median of the ITA angle and each distance metric. The line graph showing the correlation between accuracy and distance tends to jump due to the excessive influence of a single sample in bins with an extremely small number of samples. The correlation becomes more stable, and accuracy tends to be higher in ranges with a large amount of data. This result supports that skin tone bias is the main factor causing performance differences.

\subsection{Effectiveness of DRW in Categorical Evaluation}
\label{sec:effectiveness}

\begin{figure}
    \centering
    \includegraphics[width=1\linewidth]{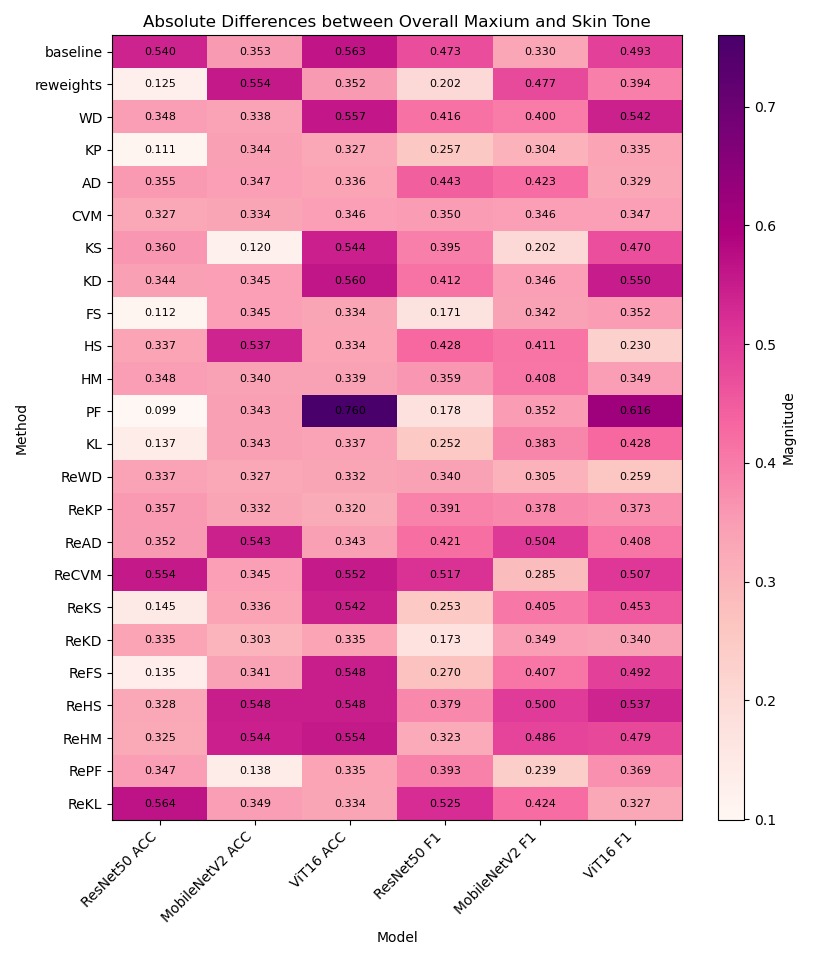}
    \Description{Description of the image for accessibility.}
    \caption{Absolute Differences between Overall Max and Skin Color Types}
    \label{fig:abs_diff_max}
\end{figure}

\begin{figure}
    \centering
    \includegraphics[width=1\linewidth]{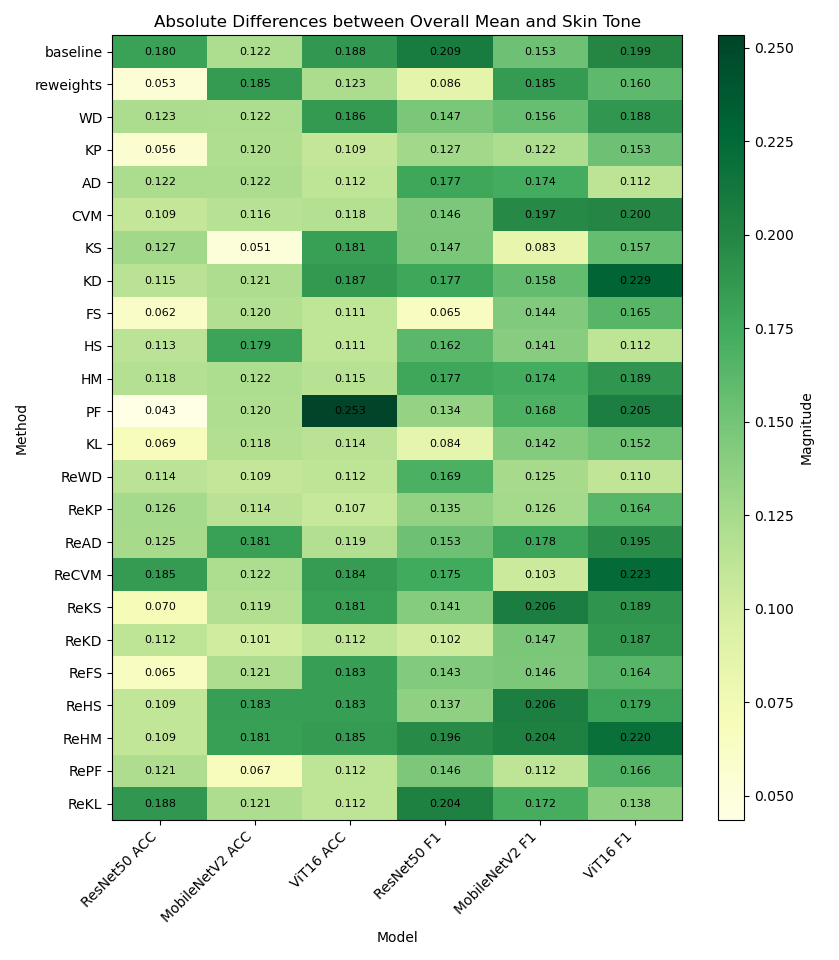}
    \Description{Description of the image for accessibility.}
    \caption{Absolute Differences between Overall Mean and Skin Color Types}
    \label{fig:abs_diff_mean}
\end{figure}

Heatmaps \ref{fig:abs_diff_max} and \ref{fig:abs_diff_mean} depict the performance differences between skin tone types when applying DRW and CARW+DRW. Heatmap \ref{fig:abs_diff_max} shows the maximum absolute values of type differences calculated from the average values of accuracy and F1 scores for each type, represented by the magnitudes of colors. In the BL, there were large differences between skin tone types, but DRW mitigated the differences in accuracy except for a few distance metrics, and group bias mitigation was confirmed when evaluated by skin tone type. Overall, it validated a better trend than CARW, which was categorical reweighting. Furthermore, the F1 score also showed a reduction in the maximum performance difference and reached a level comparable to CARW. Contrary, the results of CARW+DRW were effective in ResNet, but in ViT, improvements in both accuracy and F1 were limited for many skin tone measurement metrics.

Heatmap \ref{fig:abs_diff_mean} shows the average accuracy and F1 score of BL, as well as the sum of the absolute differences between each skin color type score. The smaller the sum of these average differences, the smaller the score differences between skin tones. First, relatively large differences were observed in both accuracy and F1 score for BL. In CARW, both accuracy and F1 score differences between skin tones were significantly reduced in ResNet, but the effect was limited in MobileNet and ViT16. In DRW, although there were differences depending on the distance method, significant improvements in accuracy were observed except for PF, and overall differences between skin color groups were reduced. In terms of the F1 score, DRW could correct group disparities in ResNet and MobileNet.It was particularly effective for AD and HS, but the effect was limited overall for ViT16. Finally, when CARW and DRW were combined (CARW+DRW), performance differences due to skin color were mitigated in accuracy and F1 score for WD, KP, AD, KD, and PF. However, the effect was not as pronounced as when DRW was used alone.

\subsection{Individual Fairness Analysis and Validity of Mitigation Methods}
\label{sec:individual}

\textbf{Baseline} Table \ref{tab:individual_correlation} presents the results of evaluating the correlation between accuracy and individual skin tone in the three models using Spearman's rank correlation coefficient. Skin tone in each image was measured comprehensively using 11 distance metrics, and correlations were calculated for each metric. In the BL, correlations were observed in most measurements except HM and PF in ResNet, with moderate correlations in WD, AD, and KD, and strong correlations in KS, KP, and KL. In MobileNet, moderate correlations were observed in KS, KP, and PF, but overall correlations were small. In ViT16, strong correlations were shown in KS, KP, and KL.

\textbf{CARW} In the second column of Table \ref{tab:individual_correlation}, which is CARW application, the correlations weakened in seven distance methods in ResNet, where the correlations were strong in BL, suggesting an improvement in unfairness. On the other hand, in MobileNet, where BL was relatively fair, the application of CARW worsened the correlations in nine distance measurements. In ViT16B, correlation strengthened in 6 indicators, and weak correlation newly appeared in WD, CVM, and FS, which were not presented in BL. In conclusion, CARW can mitigate strong correlation in some cases, but its application may impair individual fairness related to skin tone in models with inherently weak correlation.

\textbf{Our Approach (DRW)} Column 3 expresses the results of our proposed Distance Reweighting method (DRW). Based on the density of skin tone nuances captured in detail by each distance metric (Section \ref{sec:distance}), continuous reweighting was applied to the loss function. In ResNet, correlation was reduced; in other words, bias was mitigated for seven distance metrics.CARW tends to be overly effective in CVM and AD due to category-specific weighting, but the proposed method was able to reduce correlation without side effects. MobileNet was relatively fair at the BL stage, so there was little room for improvement, and although the effect was limited, further reduction in correlation was confirmed in FS and HS. In ViT16, correlation weakened in approximately half of the metrics.

\textbf{CARW+DRW} Column 4 shows the results after applying DRW following categorical reweighting. For all models, the metrics converged to an intermediate level between CARW and the distance method alone. However, the additional effect compared to the standalone application was small, and particularly for measurements like PF and KL, where the gradient of weights is large due to distance, the addition of distance weights on top of the strong category weights from CARW resulted in overweighting, leading to a deterioration in fairness in some cases.


\begin{table}[]
\caption{The Correlation between Accuracy and the Number of Training Data}
\label{tab:individual_correlation}
\centering
\begin{tabular}{c|rrrr|rrrr|rrrr}
\hline
\multirow{2}{*}{Measure} & \multicolumn{4}{c|}{ResNet}                                                                                                                   & \multicolumn{4}{c|}{MobileNet}                                                                                                                & \multicolumn{4}{c}{ViT}                                                                                                                      \\
                         & \multicolumn{1}{c}{BL} & \multicolumn{1}{c}{Re} & \multicolumn{1}{c}{Dist} & \multicolumn{1}{c|}{Re+Dist} & \multicolumn{1}{c}{BL} & \multicolumn{1}{c}{Re} & \multicolumn{1}{c}{Dist} & \multicolumn{1}{c|}{Re+Dis} & \multicolumn{1}{c}{BL} & \multicolumn{1}{c}{Re} & \multicolumn{1}{c}{Dist} & \multicolumn{1}{c}{Re+Dist} \\ \hline
WD  & -0.434 & -0.351  & -0.002 & -0.228 & 0.099 & -0.071 & 0.205 & -0.320 & 0.090 & -0.373  & -0.057  & -0.392  \\
CVM & 0.368  & -0.238  & 0.478  & 0.479  & 0.044 & 0.000  & 0.278 & 0.135  & 0.201 & 0.440   & 0.440   & 0.512   \\
AD & 0.525   & -0.205  & 0.309  & 0.287  & 0.011 & 0.283  & 0.261 & 0.289 & 0.429  & 0.505   & 0.549   & 0.162  \\
KS  & 0.851  & 0.433   & 0.459  & 0.846  & 0.486 & 0.855  & 0.882 & 0.670 & 0.812  & 0.895   & 0.947   & 0.737  \\
KP & 0.838   & 0.512   & 0.881  & 0.829  & 0.497 & 0.785 & 0.842  & 0.820 & 0.873  & 0.859   & 0.947   & 0.618 \\
KD & 0.594   & -0.201  & 0.728  & 0.817  & 0.033 & 0.200 & -0.393 & 0.050 & -0.142 & 0.233   & 0.317   & 0.402 \\
FS & 0.378   & -0.463  & -0.180  & -0.480  & -0.152  & -0.351  & -0.150  & -0.130  & -0.357 & -0.435  & 0.077  & 0.061 \\
HS & 0.353   & -0.457  & -0.227  & -0.379  & -0.166  & -0.357  & -0.155  & -0.133  & -0.390  & -0.446  & -0.407 & -0.044  \\
HM & 0.126    & -0.251  & -0.050  & -0.209 & 0.075 & 0.226 & 0.377 & 0.259   & -0.243 & -0.226  & -0.075  & -0.276 \\
PF & -0.029  & -0.696 & -0.759 & -0.759 & -0.580 & -0.638 & -0.580 & -0.638  & -0.698  & -0.203 & -0.029  & -0.029  \\
KL & 0.949  & -0.200  & 0.400 & 0.738 & -0.200 & -0.200  & -0.738 & 0.738  & -0.738  & 0.400 & 0.400 & 0.738  \\ \hline \hline
\end{tabular}
\end{table}

\begin{figure}
    \centering
    \includegraphics[width=1\linewidth]{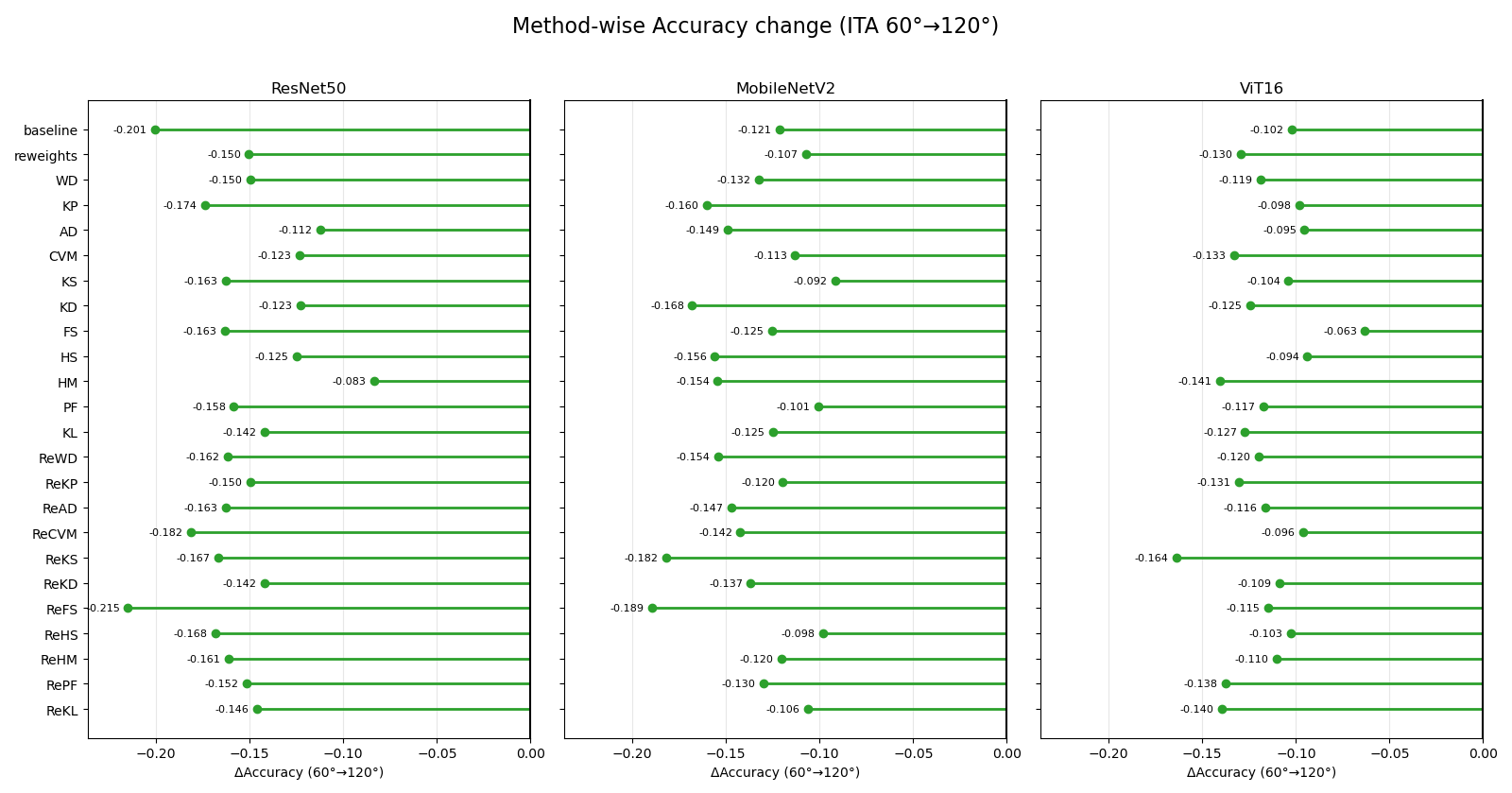}
    \Description{Description of the image for accessibility.}
    \caption{The Predicted Accuracy Change from ITA angle 60 to 120 in Skin Color Type 1}
    \label{fig:tone1_lolipop}
\end{figure}

Figure \ref{fig:tone1_lolipop} plots the accuracy change rate within skin tone type 1, which is the major skin tone type, as an indicator to confirm whether local skin tone dependency has been mitigated. This figure shows the percentage of accuracy degradation when the ITA angle value changes from 60 to 120. In the BL model, accuracy degradation was estimated to be approximately 20\% for ResNet, 12\% for MobileNet, and 10\% for ViT16. ResNet corrected local performance differences in all methods except ReFS. In MobileNet, mitigation of accuracy degradation was confirmed only in CARW, CVM, KS, PF, ReKP, ReHS, ReHM, and ReKL. In ViT16, the impact of skin tone on accuracy was suppressed only in AD, KP, FS, HM, and ReCVM.

\section{Discussion and Future Work}
\subsection{Discussion}
Let us now revisit our four research questions in light of our experimental findings.

\textbf{RQ1: Can fairness analysis based on skin types consistently detect unfairness across diverse individual skin colors?}
Our experiments confirm that categorical group-based fairness assessments, such as those derived from Fitzpatrick skin types, are insufficient to capture inequities across individuals.  Although type 1 (light skin tones - see Section \ref{sec:categorical}), skin color type 1, which was the majority, showed average accuracy and the highest performance in terms of F1 score. However, as shown in \ref{sec:tone1}), This indicates that individuals with minority tones, even within majority groups, are at risk of reduced accuracy.   These findings align with earlier concerns \cite{ guo2022bias,bevan2022detecting} that categorical fairness can mask intra-group inequities. The implication is clear: fairness evaluations for continuous and multidimensional attributes such as skin tone must move beyond group averages and incorporate distribution-aware methodologies capable of detecting subtle variations.

\textbf{RQ2: Are there correlations between the number of data instances in diverse skin colors and accuracy?} 
Our results showed consistent correlations between the density of training samples and classification accuracy across most statistical distance functions  (Section \ref{sec:correlation}). This relationship reflects a well-established principle of machine learning, that performance improves with sufficient representation, but in this context it has fairness consequences. Underrepresented tones, especially at the darker and lighter extremes, were systematically associated with lower predictive performance. The correlation strength varied depending on the chosen distance metric and model architecture, underscoring that fairness evaluations are sensitive both to methodological design and model bias. These results provide quantitative evidence for the hypothesis that disparities in dermatological AI stem primarily from imbalances in training distributions, not inherent model limitations.

\textbf{RQ3: Can reweighting using these relationships address the bias caused by minority skin tone nuances?} 
The proposed Distance Reweighting (DRW) method demonstrated consistent reductions in correlation between training sample density and accuracy, outperforming conventional categorical reweighting (CARW) in most cases  (Sections \ref{sec:effectiveness} and \ref{sec:individual}). While CARW occasionally mitigated disparities at the group level, it sometimes worsened fairness at the individual level, particularly in MobileNet where baseline correlations were already weak. DRW, by contrast, provided a more stable correction by operating directly on the continuum of skin tone distributions. These findings confirm that fairness interventions must explicitly address the continuous nature of skin tone rather than relying on discrete categories. Furthermore, DRW proved effective without to be effective without introducing additional data augmentation, which is computationally expensive.

\textbf{RQ4: Which method is most suitable to measure skin tone bias?} 
Among the twelve statistical distance functions evaluated, Fidelity Similarity (FS) consistently emerged as the most effective metric for detecting nuanced disparities and supporting bias mitigation. FS proved robust across CNN- and Transformer-based models, likely due to its ability to preserve distributional similarities while accounting for subtle tonal variations. Wasserstein Distance (WD), Hellinger Metric (HM), and Harmonic Mean Similarity (HS) also provided reliable performance, whereas purely geometric metrics such as Patrick–Fisher distance were less effective. This suggests that fairness analysis benefits from similarity- and information-theoretic measures that capture both local and global distributional properties These results extend prior work  \cite{paxton2024measuring, paxton2025evaluating} that was limited to single distance measures, and establish a comparative foundation for metric selection in fairness research.

\textbf{A Note on Architectural Sensitivity}
The impact of fairness interventions was not uniform across architectures. Convolutional models (ResNet50, MobileNetV2) exhibited clear improvements under DRW, while the Vision Transformer displayed more modest gains. This divergence suggests that CNNs, which rely heavily on local pixel distributions, are more vulnerable to imbalances in skin tone density, whereas Transformers may partially compensate through patch-based representations. These insights indicate that fairness interventions should consider not only data distributions but also model-specific inductive biases.

\subsection{Limitations}
Several limitations must be acknowledged. First, the HAM10000 dataset has been digitally preprocessed using a neural network model to enhance lesion visibility by adjusting brightness and hue, which may have distorted true skin tone distributions. Although our methods rely on relative rather than absolute differences, replication on raw, clinically diverse datasets is required for validation. Second, while we compared a wide set of distance metrics, the optimal choice may vary by context, and combining multiple measures could yield richer fairness representations. Finally, our experiments focused on skin lesion classification; further studies are needed to assess generalisability across other dermatological and medical imaging tasks.

\subsection{Implications and Future Work}
This work contributes to a growing body of research calling for fairness frameworks that move beyond categorical proxies. By modelling sensitive attributes as continuous distributions, our methodology enables a more nuanced and clinically relevant evaluation of fairness. Future research should extend this approach to other continuous sensitive variables, such as age-related imaging biomarkers, voice pitch in speech recognition, or tissue density in radiology. Additionally, hybrid strategies combining distributional reweighting with adversarial learning, data augmentation, or federated training could yield even greater robustness. Finally, collaboration with dermatologists to define clinically meaningful thresholds of tonal variation will be essential to ensure that fairness metrics align with medical practice.

\section{Conclusion}
In this study, we addressed the critical issue of fairness in dermatological AI by moving beyond categorical proxies of skin tone and instead operationalizing fairness at the individual level. Unlike prior approaches that rely on discrete classifications such as ethnicity or Fitzpatrick skin types, our methodology treated skin tone as a continuous and multidimensional attribute, thereby capturing the subtleties that group-based analyses often overlook. To this end, we systematically evaluated twelve statistical distance metrics for quantifying skin tone nuances, and proposed a novel distribution-based reweighting (DRW) framework to mitigate the uneven representation of minority tones.
Our large-scale experiments across three architectures (CNN-based models ResNet50 and MobileNetV2 and Vision Transformer Model) under 72 experimental conditions (baselines, categorical reweighting, distance reweighting, and their combinations) yielded several key insights. First, categorical fairness interventions were able to reduce disparities at a coarse level but frequently failed to address within-group inequities, sometimes even exacerbating individual-level unfairness. Second, the DRW approach consistently reduced correlations between training data density and classification accuracy, demonstrating its capacity to uncover and mitigate subtle skin tone biases. Third, among the distance metrics, Fidelity Similarity (FS) exhibited the most consistent and robust performance, though Wasserstein Distance (WD), Hellinger Metric (HM), and Harmonic Mean Similarity (HS) were proven effective in balancing fairness with accuracy. These findings confirm that fairness in continuous sensitive attributes cannot be reliably assessed, or remediated, through categorical proxies alone.

Importantly, our results also revealed architectural sensitivities: while convolutional models (ResNet50, MobileNetV2) benefitted substantially from DRW, improvements were more modest in the Vision Transformer, suggesting that the relationship between model type and fairness intervention warrants further investigation. This nuance underscores the need for fairness strategies tailored not only to data distributions but also to the inductive biases of underlying model architectures.

Beyond the specific domain of skin lesion classification, the methodological contributions of this work carry broader implications. The proposed framework is generalizable to other medical imaging tasks where continuous attributes, such as tissue density, age-related changes, or voice pitch in multimodal data, defy categorical reduction. By establishing a principled approach for measuring and correcting bias in such settings, this work provides a foundation for advancing the state of “individual fairness” in machine learning.

In conclusion, this study demonstrates that fairness in dermatological AI requires moving from group-level analyses to individual-level considerations. By integrating continuous statistical measures with reweighting strategies, we offer a pathway to more equitable, accurate, and trustworthy medical AI systems. The proposed methodology thus contributes both to the technical development of fairness-aware learning and to the ethical imperative of ensuring that diagnostic technologies serve all individuals equitably, regardless of skin tone.

\bibliographystyle{ACM-Reference-Format}
\bibliography{bibfile}

\appendix
\label{sec:appendix}

\begin{figure}
    \centering
    \includegraphics[width=1\linewidth]{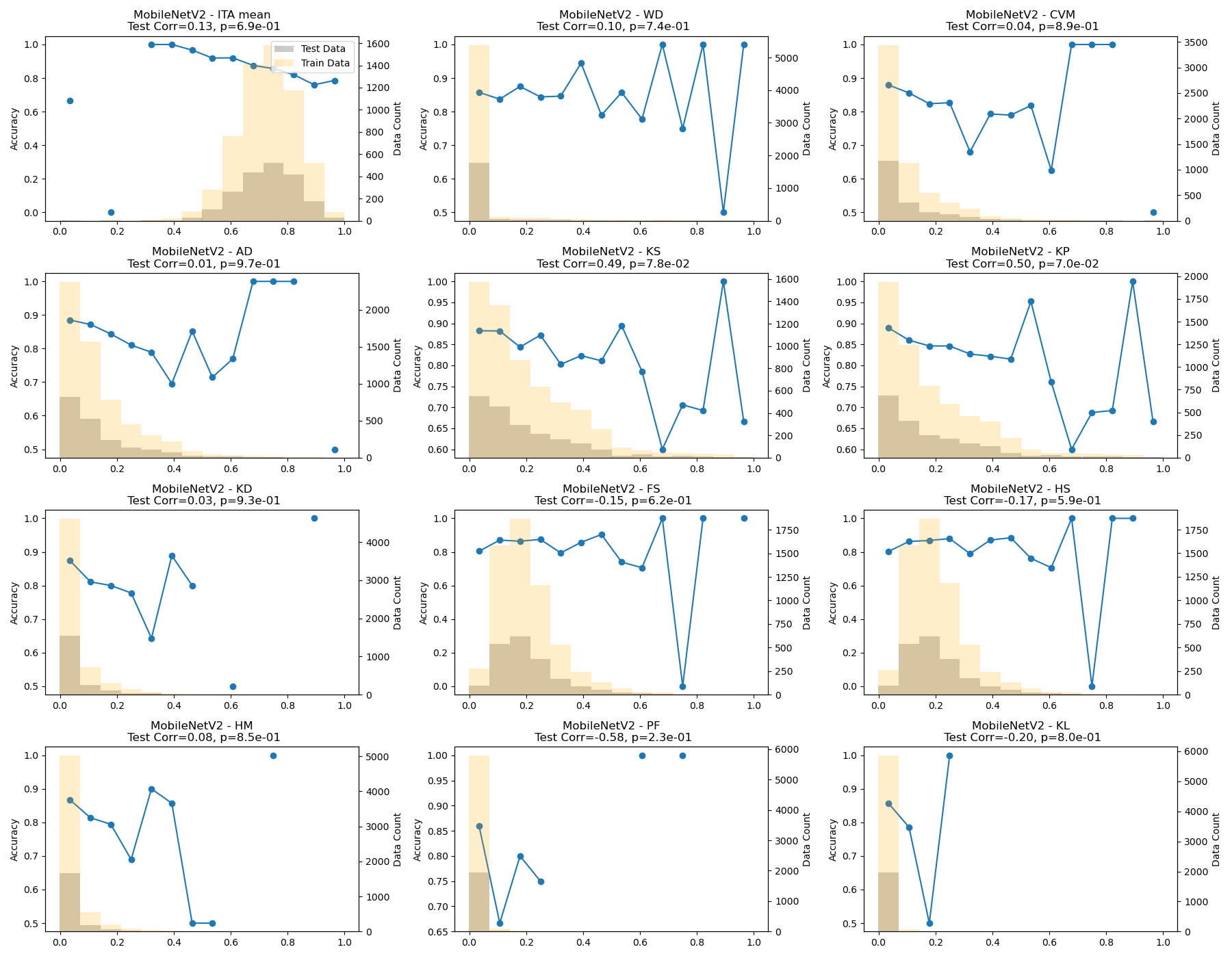}
    \Description{Description of the image for accessibility.}
    \caption{Correlation between Performance and the Number of Training Data (MobileNet)}
    \label{fig:mobilenet_individual_correlation}
\end{figure}

\begin{figure}
    \centering
    \includegraphics[width=1\linewidth]{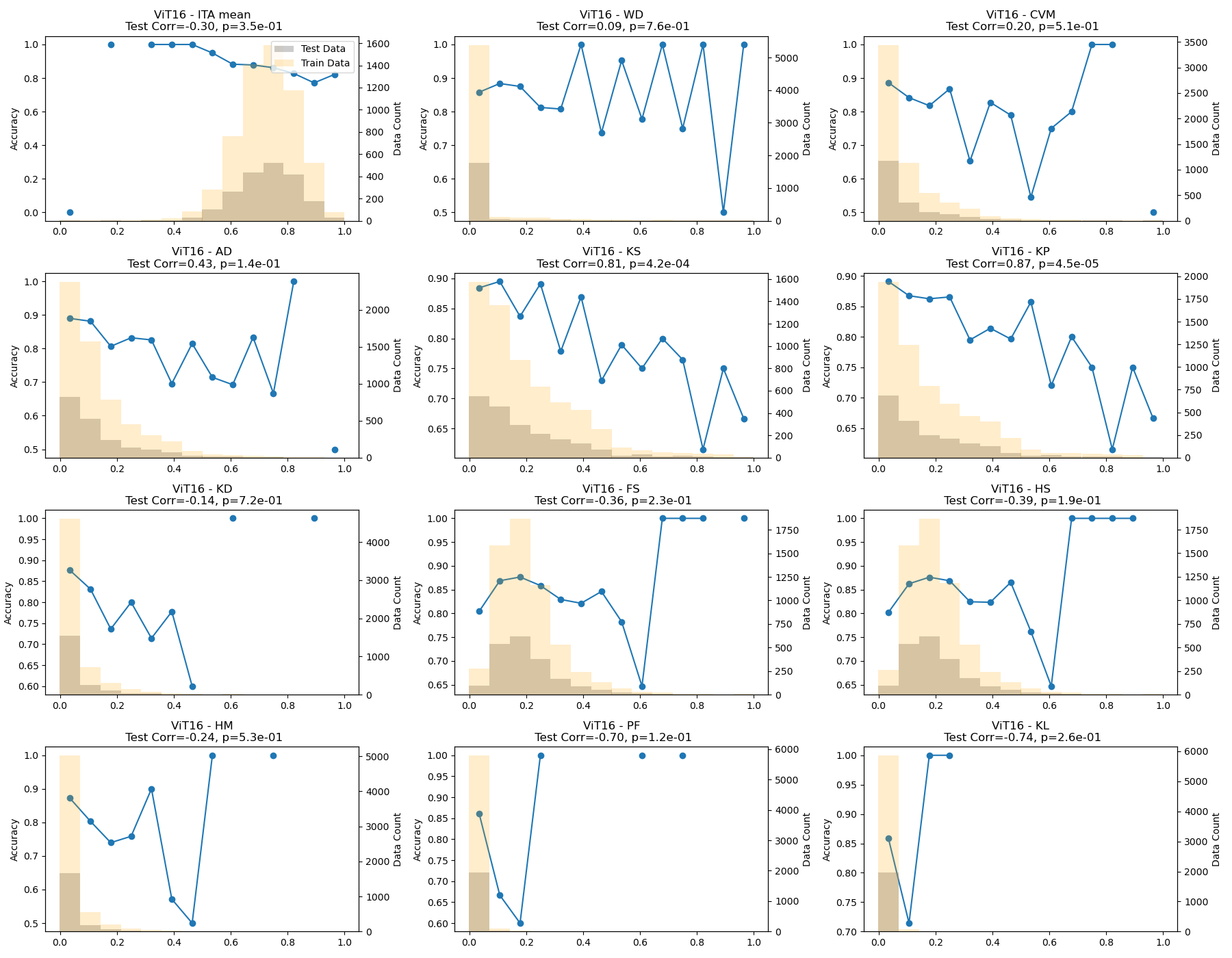}
    \Description{Description of the image for accessibility.}
    \caption{Correlation between Performance and the Number of Training Data (ViT16)}
    \label{fig:vit_individual_correlation}
\end{figure}

\begin{center}
\begin{longtable}{lll|ccc|ccc}
\caption{Long Table} \\
\toprule
\multirow{2}{*}{Method} & \multirow{2}{*}{Skin Tone} & \multirow{2}{*}{Data} & \multicolumn{3}{l}{ACC} & \multicolumn{3}{l}{F1-score}   \\
& & & ResNet50 & MobileNetV2 & ViT16 & ResNet50 & MobileNetV2 & ViT16 \\
\midrule
\endfirsthead

\toprule
\multirow{2}{*}{Method} & \multirow{2}{*}{Skin Tone} & \multirow{2}{*}{Data}
& \multicolumn{3}{c}{ACC} & \multicolumn{3}{c}{F1-score} \\
& & & ResNet50 & MobileNetV2 & ViT16 & ResNet50 & MobileNetV2 & ViT16 \\
\midrule
\endhead

\hline \hline
\endlastfoot
                             
\multirow{7}{*}{Baseline} & 1 & 1585 & 0.837 & 0.840 & 0.846 & 0.765 & 0.767 & 0.749 \\
                        & 2  & 223 & 0.928    & 0.919       & 0.883 & 0.743    & 0.718       & 0.608 \\
                        & 3  & 100 & 0.900    & 0.930       & 0.950 & 0.454    & 0.593       & 0.676 \\
                        & 4  & 56  & 0.911    & 0.929       & 0.946 & 0.878    & 0.850       & 0.927 \\
                        & 5  & 12  & 0.917    & 1.000       & 1.000 & 0.667    & 1.000       & 1.000 \\
                        & 6  & 4   & 0.250    & 0.500       & 0.250 & 0.133    & 0.389       & 0.200 \\
                        & Mean  & 330  & 0.790    & 0.853       & 0.813 & 0.606    & 0.719       & 0.693 \\ \hline
\multirow{7}{*}{Reweighting} & 1                          & 1585                  & 0.842    & 0.840       & 0.836 & 0.738    & 0.738       & 0.718 \\
                             & 2                          & 223                   & 0.928    & 0.910       & 0.901 & 0.721    & 0.680       & 0.649 \\
                             & 3                          & 100                   & 0.940    & 0.910       & 0.930 & 0.823    & 0.566       & 0.735 \\
                             & 4                          & 56                    & 0.875    & 0.911       & 0.946 & 0.773    & 0.711       & 0.927 \\
                             & 5                          & 12                    & 0.917    & 1.000       & 1.000 & 0.933    & 1.000       & 1.000 \\
                             & 6                          & 4                     & 0.750    & 0.250       & 0.500 & 0.556    & 0.167       & 0.333 \\
                             & Mean      & 330                   & 0.875    & 0.804       & 0.852 & 0.757    & 0.644       & 0.727 \\
\multirow{7}{*}{WD} & 1 & 1585                  & 0.828    & 0.811       & 0.845 & 0.725    & 0.735       & 0.769 \\
                  & 2  & 223                   & 0.946    & 0.924       & 0.897 & 0.758    & 0.732       & 0.674 \\
                  & 3  & 100                   & 0.940    & 0.920       & 0.940 & 0.856    & 0.607       & 0.656 \\
                  & 4  & 56                    & 0.875    & 0.875       & 0.911 & 0.826    & 0.658       & 0.822 \\
                  & 5  & 12                    & 1.000    & 1.000       & 1.000 & 1.000    & 1.000       & 1.000 \\
                  & 6  & 4                     & 0.500    & 0.500       & 0.250 & 0.333    & 0.267       & 0.133 \\
                 & Mean &  330                     & 0.848    & 0.838       & 0.807 & 0.750    & 0.667       & 0.676 \\ \hline
\multirow{7}{*}{KP} & 1  & 1585                  & 0.831    & 0.827       & 0.830 & 0.749    & 0.748       & 0.731 \\
                    & 2  & 223                   & 0.933    & 0.924       & 0.892 & 0.733    & 0.689       & 0.662 \\
                    & 3  & 100                   & 0.890    & 0.900       & 0.910 & 0.407    & 0.535       & 0.477 \\
                    & 4  & 56                    & 0.929    & 0.911       & 0.911 & 0.891    & 0.706       & 0.741 \\
                    & 5  & 12                    & 0.833    & 1.000       & 0.917 & 0.647    & 1.000       & 0.733 \\
                    & 6  & 4                     & 0.750    & 0.500       & 0.500 & 0.556    & 0.500       & 0.267 \\
                   & Mean &  330                     & 0.861    & 0.844       & 0.827 & 0.664    & 0.696       & 0.602 \\ \hline
\multirow{7}{*}{AD} & 1  & 1585 & 0.843 & 0.826 & 0.845 & 0.753    & 0.756       & 0.762 \\
                & 2 & 223                   & 0.919    & 0.924       & 0.892 & 0.693    & 0.702       & 0.655 \\
                & 3 & 100                   & 0.940    & 0.920       & 0.950 & 0.637    & 0.591       & 0.671 \\
                & 4  & 56                    & 0.929    & 0.911       & 0.911 & 0.906    & 0.822       & 0.822 \\
                & 5  & 12                    & 1.000    & 1.000       & 0.917 & 1.000    & 1.000       & 0.733 \\
                & 6  & 4                     & 0.500    & 0.500       & 0.500 & 0.267    & 0.267       & 0.333 \\
              & Mean &  330                     & 0.855    & 0.847       & 0.836 & 0.709    & 0.690       & 0.663 \\ \hline
\multirow{7}{*}{CVM} & 1 & 1585 & 0.830    & 0.821       & 0.838 & 0.715    & 0.739       & 0.748 \\ 
                    & 2  & 223  & 0.919    & 0.892       & 0.897 & 0.696    & 0.641       & 0.636 \\
                    & 3  & 100  & 0.920    & 0.880       & 0.910 & 0.529    & 0.470       & 0.472 \\
                    & 4  & 56  & 0.875    & 0.911       & 0.929 & 0.829    & 0.890       & 0.891 \\
                    & 5  & 12                    & 0.917    & 1.000       & 1.000 & 0.667    & 1.000       & 1.000 \\
                    & 6  & 4    & 0.500    & 0.500       & 0.500 & 0.267    & 0.333       & 0.333 \\
                    & Mean  &   330 & 0.827    & 0.834       & 0.846 & 0.617    & 0.679       & 0.680 \\ \hline
\multirow{7}{*}{KS}          & 1  & 1585 & 0.838    & 0.838       & 0.847 & 0.727    & 0.744       & 0.749 \\
                             & 2   & 223 & 0.933    & 0.906       & 0.901 & 0.734    & 0.682       & 0.656 \\
                             & 3   & 100 & 0.960    & 0.900       & 0.940 & 0.684    & 0.527       & 0.703 \\
                             & 4   & 56  & 0.929    & 0.911       & 0.911 & 0.891    & 0.822       & 0.878 \\
                             & 5   & 12  & 1.000    & 0.917       & 0.917 & 1.000    & 0.867       & 0.667 \\
                             & 6   & 4   & 0.500    & 0.750       & 0.250 & 0.333    & 0.733       & 0.167 \\
                             & Mean &  330   0.860    & 0.870       & 0.794 & 0.728    & 0.729       & 0.637 \\ \hline
\multirow{7}{*}{KD}          & 1   & 1585 & 0.848    & 0.827       & 0.842 & 0.751    & 0.733       & 0.748 \\
                             & 2   & 223  & 0.928    & 0.919       & 0.897 & 0.721    & 0.700       & 0.782 \\
                             & 3   & 100 & 0.910    & 0.930       & 0.940 & 0.559    & 0.756       & 0.548 \\
                             & 4   & 56  & 0.875    & 0.893       & 0.929 & 0.773    & 0.551       & 0.891 \\
                             & 5   & 12  & 1.000    & 1.000       & 1.000 & 1.000    & 1.000       & 1.000 \\
                             & 6   & 4   & 0.500    & 0.500       & 0.250 & 0.267    & 0.333       & 0.133 \\
                             & Mean  & 330  & 0.844    & 0.845       & 0.810 & 0.678    & 0.679       & 0.684 \\ \hline
\multirow{7}{*}{FS}          & 1  & 1585 & 0.817    & 0.830       & 0.847 & 0.726    & 0.750       & 0.759 \\
                             & 2  & 223  & 0.933    & 0.915       & 0.892 & 0.746    & 0.692       & 0.599 \\
                             & 3  & 100  & 0.930    & 0.930       & 0.920 & 0.704    & 0.586       & 0.497 \\
                             & 4  & 56   & 0.911    & 0.893       & 0.929 & 0.822    & 0.690       & 0.858 \\
                             & 5  & 12  & 0.833    & 1.000       & 0.917 & 0.806    & 1.000       & 0.733 \\
                             & 6  & 4  & 0.750    & 0.500       & 0.500 & 0.556    & 0.333       & 0.267 \\
                             & Mean  &  330 & 0.862    & 0.845       & 0.834 & 0.726    & 0.675       & 0.619 \\ \hline
\multirow{7}{*}{HS}          & 1  & 1585 & 0.833    & 0.813  & 0.849 & 0.732    & 0.720       & 0.771 \\
                             & 2 & 223 & 0.937    & 0.933       & 0.892 & 0.746    & 0.739       & 0.660 \\
                             & 3 & 100 & 0.940    & 0.920       & 0.920 & 0.639    & 0.608       & 0.514 \\
                             & 4  & 56  & 0.893    & 0.893       & 0.929 & 0.854    & 0.567       & 0.713 \\
                             & 5 & 12 & 0.917    & 0.917       & 0.917 & 0.933    & 0.667       & 0.667 \\
                             & 6  & 4  & 0.500    & 0.250       & 0.500 & 0.267    & 0.167       & 0.389 \\
                             & Mean &  330 & 0.837    & 0.787       & 0.834 & 0.695    & 0.578       & 0.619 \\ \hline
\multirow{7}{*}{HM}          & 1  & 1585 & 0.842    & 0.816       & 0.833 & 0.755    & 0.712       & 0.750 \\
                             & 2  & 223 & 0.933    & 0.915       & 0.874 & 0.743    & 0.687       & 0.575 \\
                             & 3  & 100 & 0.920    & 0.900       & 0.900 & 0.521    & 0.563       & 0.478 \\
                             & 4  & 56 & 0.893    & 0.911       & 0.929 & 0.802    & 0.822       & 0.770 \\
                             & 5  & 12 & 1.000    & 1.000       & 1.000 & 1.000    & 1.000       & 1.000 \\
                             & 6   & 4  & 0.500    & 0.500       & 0.500 & 0.333    & 0.267       & 0.333 \\
                             & Mean & 330 & 0.848    & 0.840       & 0.839 & 0.692    & 0.675       & 0.651 \\ \hline
\multirow{7}{*}{PF}          & 1  & 1585 & 0.835    & 0.826       & 0.842 & 0.754    & 0.739       & 0.751 \\
                             & 2  & 223  & 0.915    & 0.910       & 0.915 & 0.679    & 0.669       & 0.675 \\
                             & 3  & 100 & 0.930    & 0.910       & 0.940 & 0.651    & 0.550       & 0.631 \\
                             & 4  & 56 & 0.929    & 0.911       & 0.946 & 0.891    & 0.822       & 0.927 \\
                             & 5  & 12 & 1.000    & 1.000       & 0.917 & 1.000    & 1.000       & 0.714 \\
                             & 6  & 4 & 1.000    & 0.500       & 0.000 & 1.000    & 0.333       & 0.000 \\
                             & Mean & 330 & 0.935    & 0.843       & 0.760 & 0.829    & 0.685       & 0.616 \\ \hline
\multirow{7}{*}{KL}          & 1  & 1585 & 0.832    & 0.832       & 0.832 & 0.720    & 0.760       & 0.734 \\
                             & 2  & 223  & 0.928    & 0.915       & 0.879 & 0.727    & 0.715       & 0.755 \\
                             & 3  & 100  & 0.940    & 0.920       & 0.950 & 0.563    & 0.817       & 0.820 \\
                             & 4  & 56   & 0.875    & 0.893       & 0.946 & 0.742    & 0.675       & 0.927 \\
                             & 5  & 12   & 1.000    & 1.000       & 0.917 & 1.000    & 1.000       & 0.667 \\
                             & 6  & 4    & 0.750    & 0.500       & 0.500 & 0.733    & 0.333       & 0.267 \\
                             & Mean & 330 & 0.887    & 0.843       & 0.837 & 0.748    & 0.717       & 0.695 \\ \hline
\multirow{7}{*}{ReWD}        & 1 & 1585 & 0.833    & 0.829       & 0.829 & 0.723    & 0.744       & 0.739 \\
                             & 2 & 223  & 0.924    & 0.906       & 0.879 & 0.721    & 0.713       & 0.580 \\
                             & 3 & 100  & 0.920    & 0.900       & 0.920 & 0.505    & 0.570       & 0.592 \\
                             & 4 & 56   & 0.929    & 0.911       & 0.946 & 0.891    & 0.713       & 0.927 \\
                             & 5 & 12   & 0.917    & 0.917       & 0.917 & 0.867    & 0.760       & 0.667 \\
                             & 6 & 4  & 0.500    & 0.500       & 0.500 & 0.333    & 0.333       & 0.500 \\
                             & Mean & 330 & 0.837    & 0.827       & 0.832 & 0.673    & 0.639       & 0.667 \\ \hline
\multirow{7}{*}{ReKP}        & 1  & 1585 & 0.836    & 0.823       & 0.834 & 0.734    & 0.737       & 0.730 \\
                             & 2  & 223  & 0.933    & 0.915       & 0.897 & 0.744    & 0.681       & 0.780 \\
                             & 3  & 100  & 0.960    & 0.930       & 0.930 & 0.710    & 0.807       & 0.521 \\
                             & 4  & 56   & 0.911    & 0.911       & 0.929 & 0.822    & 0.711       & 0.891 \\
                             & 5  & 12   & 1.000    & 0.917       & 0.833 & 1.000    & 0.667       & 0.647 \\
                             & 6  & 4    & 0.500    & 0.500       & 0.500 & 0.333    & 0.267       & 0.267 \\
                             & Mean & 330 & 0.857    & 0.832       & 0.820 & 0.724    & 0.645       & 0.639 \\ \hline
\multirow{7}{*}{ReAD}        & 1  & 1585  & 0.831    & 0.825       & 0.830 & 0.736    & 0.750       & 0.733 \\
                             & 2  & 223   & 0.933    & 0.915       & 0.897 & 0.733    & 0.695       & 0.681 \\
                             & 3  & 100  & 0.940    & 0.910       & 0.920 & 0.859    & 0.775       & 0.498 \\
                             & 4  & 56   & 0.911    & 0.857       & 0.911 & 0.861    & 0.641       & 0.871 \\
                             & 5  & 12   & 1.000    & 1.000       & 1.000 & 1.000    & 1.000       & 1.000 \\
                             & 6  & 4    & 0.500    & 0.250       & 0.500 & 0.333    & 0.167       & 0.267 \\
                             & Mean & 330 & 0.852    & 0.793       & 0.843 & 0.754    & 0.671       & 0.675 \\ \hline
\multirow{7}{*}{ReCVM}       & 1  & 1585 & 0.834    & 0.824       & 0.833 & 0.751    & 0.727       & 0.742 \\
                             & 2  & 223  & 0.919    & 0.933       & 0.892 & 0.710    & 0.727       & 0.763 \\
                             & 3  & 100  & 0.910    & 0.920       & 0.910 & 0.753    & 0.626       & 0.510 \\
                             & 4  & 56   & 0.911    & 0.893       & 0.929 & 0.889    & 0.709       & 0.858 \\
                             & 5  & 12   & 1.000    & 1.000       & 1.000 & 1.000    & 1.000       & 1.000 \\
                             & 6  & 4   & 0.250    & 0.500       & 0.250 & 0.200    & 0.500       & 0.167 \\
                             & Mean & 330 & 0.804    & 0.845       & 0.802 & 0.717    & 0.715       & 0.673 \\ \hline
\multirow{7}{*}{ReKS}        & 1  & 1585 & 0.831    & 0.815       & 0.831 & 0.742    & 0.743       & 0.740 \\
                             & 2  & 223  & 0.924    & 0.924       & 0.888 & 0.702    & 0.740       & 0.649 \\
                             & 3  & 100  & 0.920    & 0.870       & 0.920 & 0.564    & 0.457       & 0.505 \\
                             & 4  & 56  & 0.946    & 0.911       & 0.946 & 0.919    & 0.822       & 0.927 \\
                             & 5  & 12  & 1.000    & 1.000       & 0.917 & 1.000    & 1.000       & 0.733 \\
                             & 6  & 4  & 0.750    & 0.500       & 0.250 & 0.556    & 0.267       & 0.167 \\
                             & Mean & 0.895    & 0.836       & 0.792 & 0.747    & 0.671       & 0.620 \\ \hline
\multirow{7}{*}{ReKD}        & 1  & 1585 & 0.838    & 0.817       & 0.841 & 0.722    & 0.723       & 0.755 \\
                             & 2  & 223 & 0.933    & 0.906       & 0.883 & 0.722    & 0.675       & 0.618 \\
                             & 3  & 100 & 0.910    & 0.920       & 0.920 & 0.538    & 0.522       & 0.508 \\
                             & 4  & 56  & 0.911    & 0.839       & 0.946 & 0.822    & 0.667       & 0.927 \\
                             & 5  & 12  & 0.917    & 0.833       & 0.917 & 0.733    & 0.839       & 0.897 \\
                             & 6  & 4  & 0.500    & 0.500       & 0.500 & 0.500    & 0.267       & 0.333 \\
                             & Mean &  330  & 0.835    & 0.803       & 0.835 & 0.673    & 0.615       & 0.673 \\ \hline
\multirow{7}{*}{ReFS}        & 1  & 1585  & 0.825    & 0.819       & 0.849 & 0.736    & 0.752       & 0.770 \\
                             & 2  & 223   & 0.915    & 0.919       & 0.892 & 0.725    & 0.706       & 0.660 \\
                             & 3  & 100   & 0.910    & 0.930       & 0.950 & 0.481    & 0.642       & 0.685 \\
                             & 4  & 56    & 0.911    & 0.875       & 0.929 & 0.882    & 0.673       & 0.770 \\
                             & 5   & 12  & 1.000    & 1.000       & 0.917 & 1.000    & 1.000       & 0.733 \\
                             & 6   & 4   & 0.750    & 0.500       & 0.250 & 0.556    & 0.267       & 0.133 \\
                             & Mean &  330 & 0.885    & 0.841       & 0.798 & 0.730    & 0.673       & 0.625 \\ \hline
\multirow{7}{*}{ReHS}        & 1  & 1585 & 0.830    & 0.824       & 0.847 & 0.725    & 0.739       & 0.766 \\
                             & 2  & 223  & 0.901    & 0.919       & 0.910 & 0.667    & 0.716       & 0.712 \\
                             & 3  & 100  & 0.930    & 0.920       & 0.920 & 0.615    & 0.516       & 0.753 \\
                             & 4  & 56   & 0.893    & 0.875       & 0.946 & 0.842    & 0.695       & 0.927 \\
                             & 5  & 12   & 0.917    & 1.000       & 0.917 & 0.760    & 1.000       & 0.733 \\
                             & 6   & 4   & 0.500    & 0.250       & 0.250 & 0.267    & 0.133       & 0.133 \\
                             & Mean  & 330 & 0.828    & 0.798       & 0.798 & 0.646    & 0.633       & 0.671 \\ \hline
\multirow{7}{*}{ReHM}        & 1  & 1585 & 0.821    & 0.820       & 0.840 & 0.740    & 0.735       & 0.770 \\
                             & 2  & 223  & 0.928    & 0.910       & 0.897 & 0.735    & 0.675       & 0.660 \\
                             & 3  & 100  & 0.890    & 0.890       & 0.890 & 0.390    & 0.526       & 0.519 \\
                             & 4  & 56  & 0.893    & 0.893       & 0.946 & 0.842    & 0.812       & 0.927 \\
                             & 5  & 12  & 0.917    & 1.000       & 1.000 & 0.897    & 1.000       & 1.000 \\
                             & 6  & 4  & 0.500    & 0.250       & 0.250 & 0.333    & 0.167       & 0.200 \\
                             & Mean & 330 & 0.825    & 0.794       & 0.804 & 0.656    & 0.653       & 0.679 \\ \hline
\multirow{7}{*}{RePF}        & 1  & 1585 & 0.831    & 0.823       & 0.839 & 0.744    & 0.734       & 0.765 \\
                             & 2  & 223  & 0.901    & 0.915       & 0.888 & 0.651    & 0.702       & 0.624 \\
                             & 3  & 100  & 0.940    & 0.910       & 0.920 & 0.623    & 0.541       & 0.518 \\
                             & 4  & 56   & 0.911    & 0.929       & 0.946 & 0.673    & 0.858       & 0.927 \\
                             & 5  & 12   & 1.000    & 1.000       & 0.917 & 1.000    & 1.000       & 0.714 \\
                             & 6  & 4    & 0.500    & 0.750       & 0.500 & 0.267    & 0.733       & 0.267 \\
                             & Mean & 330 & 0.847    & 0.888       & 0.835 & 0.660    & 0.761       & 0.636 \\ \hline
\multirow{7}{*}{ReKL}        & 1  & 1585  & 0.842    & 0.837       & 0.832 & 0.745    & 0.740       & 0.748 \\
                             & 2  & 223 & 0.933    & 0.919       & 0.915 & 0.742    & 0.697       & 0.682 \\
                             & 3  & 100 & 0.930    & 0.930       & 0.910 & 0.604    & 0.600       & 0.573 \\
                             & 4  & 56  & 0.929    & 0.911       & 0.929 & 0.891    & 0.842       & 0.891 \\
                             & 5  & 12  & 1.000    & 1.000       & 0.917 & 1.000    & 1.000       & 0.733 \\
                             & 6   & 4  & 0.250    & 0.500       & 0.500 & 0.167    & 0.267       & 0.333 \\
                             & Mean &  330 & 0.814    & 0.849       & 0.834 & 0.692    & 0.691       & 0.660 
\end{longtable}
\end{center}

\end{document}